\pdfoutput=1

\documentclass[11pt]{article}

\usepackage[]{acl}

\usepackage{times}
\usepackage{latexsym}

\usepackage[T1]{fontenc}

\usepackage[utf8]{inputenc}

\usepackage{microtype}

\title{Enhancing LLM Intelligence with ARM-RAG:
Auxiliary Rationale Memory for Retrieval Augmented Generation}

\author{Eric Melz \\
  SearchStax \\
  \texttt{eric@emelz.com} 
}

\begin{document}
\maketitle
\begin{abstract}
Large Language Models (LLMs) are smart but forgetful.  Recent studies, (e.g., \cite{bubeck2023sparks}) on modern LLMs have shown that they are capable of performing amazing tasks typically necessitating human-level intelligence.  However, unlike humans, frozen LLMs do not improve over time; they neither acquire new knowledge nor learn from their successes or failures.

Some approaches to improving the intelligence of LLMs include fine-tuning models based on problem-solving 
performance \cite{zelikman2022star}, and building bigger and more sophisticated models \cite{bubeck2023sparks}.  
However, these methods have the drawback of requiring substantial data and computational resources to retrain existing models.

In this paper, we explore the use of Retrieval Augmented Generation, also known as RAG \cite{lewis2021retrievalaugmented} to improve problem-solving performance.
We propose ARM-RAG (Auxiliary Rationale Memory for Retrieval Augmented Generation), a system that learns from its successes without incurring high training costs.  We demonstrate that the storage and subsequent retrieval of reasoning chains have a positive influence on performance in grade-school math problems.
\end{abstract}

\section{Introduction}\label{sec:intro}

Large Language Models (LLMs) are smart but forgetful.  Recent studies, (e.g., \cite{bubeck2023sparks}) on modern LLMs have shown that they are capable of performing amazing tasks typically necessitating human-level intelligence.  However, unlike humans, frozen LLMs do not improve over time; they neither acquire new knowledge nor learn from their successes or failures.

Several approaches exist to enhance the performance of LLMs.  One effective strategy is to train larger LLMs using more data and extensive fine-tuning.  For example, \cite{bubeck2023sparks} demonstrate that GPT-4 significantly outperforms GPT-3 on a variety of challenging tasks.

Another approach to improving LLM performance involves fine-tuning a base LLM based on its successes and failures in problem-solving.  \cite{zelikman2022star} propose a system that enhances a base LLM by training it with examples generated from both successful and unsuccessful problem-solving attempts.  They demonstrate that this approach can significantly increase the "intelligence" of LLMs, enabling them to perform better on math problems.  However, they note that if the base LLM is not sufficiently advanced, their bootstrapping approach is ineffective.  For instance,  their model shows the desired improvements with GPT-3 but does not show any improvement when starting with GPT-2.

Retrieval Augmented Generation, aka RAG \cite{lewis2021retrievalaugmented}  has been proposed to augment the parametric memory of LLMs with the non-parametric memory of Knowledge Bases (KBs), which
can be retrieved using search engines.  The RAG approach has been shown to improve the performance of several tasks, such as Question Answering requiring multiple "hops", as demonstrated by \cite{DBLP:journals/corr/abs-2101-00436}.

All of the aforementioned techniques have shown improvements in performance over base LLMs.  However, each has its drawbacks.  The build-from-scratch and fine-tuning approaches require substantial amounts of data and computing resources for model training.  In the case of RAG, once the language model and the retrieval model are established, both the parametric and non-parametric memories become fixed, and no further learning occurs over time.  

The central hypothesis of this paper is that Retrieval Augmented Generation (RAG) can be
successfully deployed to enhance the problem-solving abilities of LLMs.  We propose ARM-RAG (Auxiliary Rationale Memory for Retrieval Augmented Generation), a system that learns from its successes without incurring high training costs.  ARM-RAG retains the detailed reasoning steps it took when it successfully answered problems.  We demonstrate that the retrieval of these reasoning chains, known as "Rationales", improves the performance of subsequent problem-solving attempts.

\section{Prior Literature}

In this section, we review the literature on three topics pertinent to the current work.
First, we examine the recent trend of LLMs in tackling increasingly complex problems.  Next,
we look at the emergence of Retrieval Augmented Generation (RAG) and its impact on intelligent systems.
Finally, we explore the literature on the use of chain-of-thought reasoning patterns to enhance
LLM performance in problem-solving tasks.

\subsection{LLMs are trending toward AGI}
\cite{bubeck2023sparks} claim that modern LLMs exhibit "sparks of AGI (Artificial General Intelligence)".
In their paper, they extensively study GPT-4 and argue that it is significantly 
more powerful than the previous releases of GPT, even approaching human intelligence for
some tasks.  For instance, they compare GPT-3.5 versus GPT-4 on a math problem depicted in Figure 
\ref{fig:prompt}.  The answer produced by GPT-3.5, shown in Figure \ref{fig:gpt35}
is nonsensical, while the answer produced by GPT-4, presented in Figure \ref{fig:gpt4}, is mostly coherent.

\begin{figure}[h]
\begin{small}
Within the duration of one year, a rabbit
population first multiplies itself by a
factor $a$ and on the last day of the year
$b$ rabbits are taken by humans for 
adoption.  Supposing that on first day 
of the first year there are $x$ rabbits, 
we know that exactly 3 years afterwards 
there will be a population of $27x - 26$
rabbits. What are the values of $a$ and $b$?
\end{small}
\caption{Example prompt.}
\label{fig:prompt}
\end{figure}

\begin{figure}[h]
\begin{small}
Let n be the total number of rabbits at
the end of the first year. Then,
\[
n = ax
\]  
At the end of the third year, there will
be $27x - 26$ rabbits. Therefore,
\[
  (ax - b)(a^2x - b)(a^3x - b) = 27x - 26
\]
Expanding the left side of the
equation, we get
\[
  a^6x^3 - (a^3 + a^2 + a + 1)bx^2 +
  (a^2 +a + 1)b^2x - b^3 =
  27x - 26
\]
\end{small}
\caption{GPT-3.5 Answer}
\label{fig:gpt35}
\end{figure}

\begin{figure}[h]
\begin{small}
Let’s call the rabbit population after one year $y$. We know that
$y = ax - b$.  After two years, the population will be $ay -b = a(ax - b) -b = a^2x - ab - b$.
After three years, the population will be $a(a^2x - ab - b) - b = a^3x - a^2b - ab - b$.
We are given that this is equal to $27x - 26$, so we have the equation:
\[
a^3x - a^2b - ab - b = 27x - 26
\]
We can rewrite this equation as
\[
a^3x - 27x = a^2b + ab + b - 26
\]
Now we can factor out an $x$ on the left side and a $b$ on the right side:
\[
x(a^3 - 27) = b(a^2 + a + 1 - 26/b)
\]
We can see that $a$ = 3 is a solution for the left side of the equation, so we can
substitute that in:
\[
x(0) = b(9 + 3 + 1 - 26/b)
\]
This simplified to:
\[
0 = 13b - 26
\]
Solving for $b$, we find that $b = 2$.  So $a = 3$ and $b = 2$ are the values we are looking for.
\end{small}
\caption{GPT-4 Answer}
\label{fig:gpt4}
\end{figure}

This and other examples provide compelling evidence that LLMs are rapidly
advancing in intelligence and, in some instances, can solve problems as well or better
than humans.  While this is a positive development, as mentioned in Section \ref{sec:intro},
even the most advanced LLMs are limited by the inherent fact that they possess only the
memory with which they were initially equipped and do not improve over time without
costly retraining.

\subsection{Retrieval Augmented Generation}
\cite{lewis2021retrievalaugmented} describe Retrieval Augmented Generation (RAG), which combines
a retrieval model with a language model.  
The retrieval model, referred to as the retriever or non-parametric memory, is an information retrieval system. The language model is known as the generator or parametric memory.  The retrieval model is an Information Retrieval (IR) model.  In this paper, the generator is based on FAISS \cite{faiss} dense representations and employs Maximum Inner Product Search (MIPS) to select the top-k documents from the non-parametric memory. The generator is BART \cite{lewis2019bart}, a pre-trained seq2seq transformer with 400 million parameters, trained on a diverse set of generation tasks.

Generally, tasks utilizing the RAG system proceed in two stages. First, documents are retrieved and a task-specific prompt is constructed. Next, this prompt is used as input for the generator to produce the system's response.

The RAG system has been applied to a variety of tasks, including open-domain question-answering, Jeopardy-style question generation, and fact verification. In the open-domain question-answering task, the RAG model outperforms other state-of-the-art models. The authors observe that, unlike traditional question-answering systems that depend on retrieving a passage and then extracting an answer from it, RAG systems can generate correct answers even when the answer is not explicitly present in any of the retrieved passages. This capability arises because the generator can use the input passages as cues, in conjunction with its own parametric memory, to formulate answers.

RAG-style systems have been adopted by other researchers to tackle challenging problems. For instance,  \cite{DBLP:journals/corr/abs-2101-00436}, describe Baleen, a RAG system designed for multi-hop question answering and claim verification tasks. These tasks require the extraction of evidence from two or more documents to produce a correct answer.  

Consider the claim shown in Figure \ref{fig:claim}.  Baleen's task is to determine whether the claim is supported by the evidence or not. Moreover, there is no single passage containing all the necessary evidence to substantiate the claim.

\begin{figure}[h]
\begin{small}
The MVP of a game Red Flaherty umpired was elected to the Baseball Hall of Fame.
\end{small}
\caption{Claim posed to Baleen}
\label{fig:claim}
\end{figure}

Baleen conducts multi-hop claim verification by iteratively retrieving passages and invoking the generator with the original claim and summaries of the retrieved passages. This process continues for several steps until an answer is produced or it is determined that no answer can be found. An example of these steps is illustrated in Table \ref{table:baleen}.

The retrieval model used by Baleen is FLIPR, a neural IR model similar to ColBERT \cite{khattab2020colbert}, which performs
token-level matching of queries to passages.  The generation component is ELECTRA-large \cite{ni2022electra}.

Baleen performs better than competing systems on the HoVer \cite{jiang2020hover} claim verification set, and the 
HotPotQA \cite{yang2018hotpotqa} question-answering set.

\begin{table}[h!]
    \centering
\begin{small}
    \begin{tabular}{|p{0.03\textwidth} p{0.4\textwidth}|}
    \hline
    $Q_0$ & The MVP of a game Red Flaherty umpired was elected to the Baseball Hall of Fame. \\
    \hline
    $Q_1$ & {\color{gray!80}The MVP of a game Red Flaherty umpired was elected to the Baseball Hall of Fame.}
    \textbf{Red Flaherty}:  He umpired in World Series 1955, 1958, 1965, and 1970. \\
    \hline
    $Q_2$ & {\color{gray!80}The MVP of [a] game Red Flaherty umpired was elected to the Baseball Hall of Fame. Red Flaherty: He umpired in World Series 1955, 1958, 1965, and 1970.} \textbf{1965 World Series}: It is remembered for MVP Sandy Koufax. \\
    \hline
    $Q_3$ & {\color{gray!80}The MVP of [a] game Red Flaherty umpired was elected to the Baseball Hall of Fame. Red Flaherty: He umpired in World Series 1955, 1958, 1965, and 1970. 1965 World Series: It is remembered for MVP Sandy Koufax.} \textbf{Sandy Koufax}: He was elected to the Baseball Hall of Fame.\\
    \hline
   \end{tabular}
\end{small}
\caption{Baleen reasoning hops}
\label{table:baleen}
\end{table}

\subsection{Chain-of-thought reasoning}
Human decision-making often results from extended chains of thought. It has been demonstrated that prompting with explicit intermediate reasoning can enhance the performance of language models on complex tasks.  \cite{zelikman2022star} investigate how these reasoning chains, also known as rationales, can be utilized in a feedback loop to improve the performance of LLMs. The project's goal is to enhance the quality of the rationales generated by LLMs, thereby improving the model's accuracy on problems that require reasoning.

They observe that improving rationale generation can be achieved by fine-tuning a set of rationales; however, manually creating such a dataset can be extremely labor-intensive. Their approach is to leverage the pre-existing reasoning abilities of LLMs to iteratively bootstrap a model's capacity to generate high-quality rationales. Their bootstrapping protocol unfolds in three steps. First, they prompt an LLM with a few examples to "self-generate" rationales. Next, they refine the model's ability to produce better rationales by fine-tuning it with those rationales that lead to correct answers. Finally, they repeat the process with the improved model until no further performance enhancements are observed.

They note that their bootstrapping routine enhances performance on familiar problems but falls short in solving new ones because the model does not receive feedback for incorrectly answered problems. To address this, for every problem the model fails to solve, they manually create a \textit{rationalization} that includes the correct answer and incorporate both the problem and its rationalization into the fine-tuning training set. This method improves performance on previously unseen problems.

To evaluate their system, they use three data sets.  The first is a generated data set that synthesizes
multi-digit integer addition problems.  Each example includes an input, an answer, and
a "scratchpad" that breaks down the individual steps required to solve the problem correctly.
An example of the multi-digit addition problem is depicted in Figure \ref{fig:arithmetic}.

\begin{figure}[h]
\begin{small}
\begin{verbatim}
       Input:
       6 2 4 + 2 5 9
       Target:
       <scratch>
       6 2 4 + 2 5 9 , C: 0
       2 + 5 , 3  C: 1
       6 + 2 , 8 3  C: 0
       , 8 8 3  C: 0
       0 8 8 3
       </scratch>
       8 8 3
\end{verbatim}
\end{small}
\caption{Arithmetic example.  C corresponds to the carry from the previous digit's summation.}
\label{fig:arithmetic}
\end{figure}

The next dataset they use is a common sense dataset called \textit{CommonsenseQA} \cite{talmor2019commonsenseqa}.  This data set comprises multiple-choice questions
about straightforward common-sense scenarios that necessitate world knowledge.  Answers are 
provided, along with rationales for the answers.  An example is presented in Figure \ref{fig:cqa}.

\begin{figure}[h]
\begin{small}
\begin{verbatim}
Q: Billy bought coffee and waited for 
his wife to arrive from France.  
Where might he have been?
Answer Choices:
(a) airport
(b) grocery store
(c) internet cafe
(d) supermarket
(e) train station
A: The answer must be a place where 
Billy could have been waiting for his 
wife to arrive from France. The airport 
is a place where people can wait for 
flights. Therefore, the answer is train 
station (e).
\end{verbatim}
\end{small}
\caption{CQA example.}
\label{fig:cqa}
\end{figure}

The final dataset they use is the GSM8K \cite{cobbe2021training} dataset.
This dataset contains a series of grade-school-level math problems, complete with answers and detailed reasoning steps that lead to those answers.
An example from this dataset is displayed in
in Figure \ref{fig:gsm8k}.

\begin{figure}[h]
\begin{small}
\begin{verbatim}
Q: Natalia sold clips to 48 of her 
friends in April, and then she sold 
half as many clips in May. 
How many clips did Natalia sell 
altogether in April and May?
A: Natalia sold 48/2 = <<48/2=24>>24 
clips in May.
Natalia sold 48+24 = <<48+24=72>>
72 clips altogether in April and May.
#### 72
\end{verbatim}
\end{small}
\caption{GSM8K example.}
\label{fig:gsm8k}
\end{figure}

The base model the authors use GPT-J \cite{wang2021mesh}.  On the arithmetic dataset, the baseline model achieved an accuracy of 76.3\%, and the STaR model achieved an accuracy of 89.5\% accuracy after 16 iterations.  On the CommonsenseQA dataset, they achieved an accuracy of 72.5\%, surpassing the baseline performance of 20\%.  On the GSM8K dataset, they achieved a test accuracy of only 10.7\%, and they noted that the use of rationalizations did not significantly improve system performance.

\section{Data}

To evaluate the problem-solving capabilities of our system, we utilize the GSM8K dataset.
We acquired the actual data from the GitHub repository of the STaR project \cite{STaRRepo}.  

The dataset contains 7,473 examples. We divided the dataset into two parts: 5,000 examples for training and 2,473 examples for testing.

\section{Model}

Experiments in this paper utilize a language model and, optionally, a retrieval model. The language model we employ is \texttt{gpt-3.5-turbo} \cite{gpt35_2023}.  We access the language model via the OpenAI API. The retrieval model used is \texttt{Pyserini} \cite{10.1145/3404835.3463238}, a Python 
information retrieval library that employs dense representations supported by the \texttt{Faiss} library \cite{faiss}.  

Experiments were conducted using the Google Colab environment \cite{google_colab}, with multiple notebooks 
running in parallel on Nvidia A100 GPUs.  Code for this paper's experiments can be found at \cite{arm-rag}.

\section{Experiments}
This section outlines a series of experiments conducted on the GSM8K dataset.

\subsection{Case Study - GPT-3.5 on a single example}
In this case study, we examine how GPT-3.5 (hereafter referred to as GPT) processes a single math problem. The problem in question, displayed in \ref{fig:caseStudyQuestion}, requires GPT to calculate the profit made by a house flipper after investing in a property.

\begin{figure}[h]
\begin{small}
\begin{verbatim}
Josh decides to try flipping a house.  
He buys a house for $80,000 and then puts
in $50,000 in repairs.  This increased the 
value of the house by 150%.  How much 
profit did he make?
\end{verbatim}
\end{small}
\caption{Case Study Question.}
\label{fig:caseStudyQuestion}
\end{figure}

GPT answers the question, along with a detailed step-by-step justification
for the answer.  As shown in Figure \ref{fig:goodAnswer}, GPT is capable of producing
a correct answer accompanied by clear and logical reasoning steps.

\begin{figure}[h]
\begin{small}
\begin{verbatim}
1. The increase in value was 
   80000*1.5=
   $<<80000*1.5=120000>>120,000
2. So the house is now worth 
   120000+80000=
   $<<120000+80000=200000>>200,000
3. So he made a profit of 
   200000-80000-50000=
   $<<200000-80000-50000=70000>>70,000. 
4. Answer: \\boxed{70000}.
\end{verbatim}
\end{small}
\caption{Correct Answer.  Line numbers are added for clarity.}
\label{fig:goodAnswer}
\end{figure}

GPT exhibits a degree of randomness, resulting in a variety of answers, some correct and some incorrect. Figure \ref{fig:badAnswer} presents an example of an incorrect answer. The reason for the mistake is subtle and akin to an error a human might make. A closer look at the incorrect answer reveals that the calculation of the house's initial value is flawed. In line 2, it incorrectly calculates the initial value of the house by adding the purchase price to the value of the repairs. This is a mistake because the cost of repairs should not be factored into the initial value. This miscalculation affects all subsequent steps, ultimately leading to an incorrect final answer.

\begin{figure}[h]
\begin{small}
\begin{verbatim}
1. The value of the house increased by 
   150%, so this means there was an 
   increase of 150/100 = 
   $<<150/100*80000=120000>>120,000
2. So the total value of the house was 
   80,000+50,000 = 
   $<<80000+50000=130000>>130,000
3. This means the value of the house 
   increased to 130,000+120,000 = 
   $<<130000+120000=250000>>250,000
4. His profit was 250,000-130,000 = 
   $<<250000-130000=120000>>120,000. 
5. Answer: \\boxed{120000}.
\end{verbatim}
\end{small}
\caption{Incorrect Answer.  Line numbers are added for clarity.}
\label{fig:badAnswer}
\end{figure}

The ability of GPT to produce coherent answers to math problems is noteworthy, and even the incorrect answers are intriguing because they are nearly correct. The fact that GPT generates chains of reasoning, which vary in quality, suggests the possibility that GPT could be encouraged to produce higher-quality chains of thought, and consequently, higher-quality answers. This hypothesis forms the basis of the subsequent experiments.

\subsection{Experiment 1: Single-Question performance}\label{subsec:exp1}
Given the non-deterministic nature of answers under default GPT settings, it is informative to assess how frequently a given question yields a correct answer. In this experiment, we prompted GPT 100 times with the question depicted in Figure 7 and measured the overall accuracy of the responses. Accuracy is defined as the proportion of answers that are correct.  In this case, the accuracy was 34\%.

\subsection{Experiment 2: Effect of strong prompting}\label{subsec:exp2}
The central hypothesis of this paper is that the performance of GPT can be enhanced through the use of prompting hints. The most direct hint is to provide the answer while posing the question. In this experiment, we supplied a prompt that included five examples of question/answer pairs, with each question being the one shown in Figure \ref{fig:caseStudyQuestion}, and each answer being a correct one drawn from the correct responses generated in Section \ref{subsec:exp1}. Following the same protocol as the previous experiment, we executed the query 100 times, resulting in an accuracy of 80\%. This strongly suggests that hinting can significantly influence the performance of the system. However, this experiment does not clarify whether the system has learned anything substantive. It is possible that the inclusion of certain numbers in the hints caused GPT to favor those numbers. Further research is required to ascertain the exact reasons for the improved accuracy.

\subsection{Experiment 3: Effect of strong negative prompting}\label{subsec:exp3}
Following the experiment described in Section \ref{subsec:exp2}, we further investigated the effect of hints by prompting with incorrect answers instead of correct ones. This experiment mirrored the previous one, except that the prompts were constructed using only incorrect answers derived from Experiment 1. The accuracy of this experiment was 39\%, a modest increase from the baseline accuracy of 34\%. This result suggests that while certain types of prompts can markedly enhance performance, there are other, less effective prompts that exert only a marginal influence on performance.

\subsection{Experiment 4: Baseline Training Set}\label{subsec:exp4}
In this experiment, we assessed the baseline performance of GPT by prompting it with each question in the training set exactly once. This procedure yielded a baseline accuracy of 73.2\%.

\subsection{Experiment 5: Multi-attempt questioning}\label{subsec:exp5}
Building on the findings from Section 3.1, which indicated that questions answered incorrectly might be answered correctly upon subsequent attempts, we conducted an experiment where each question in the training set was posed to GPT five times. If any of these attempts resulted in a correct answer, the question was marked as correctly answered. This approach led to an accuracy of 91.9\%, a substantial increase from the baseline accuracy of 73.2\%.

\subsection{Experiment 6: Basic ARM-RAG}\label{subsec:exp6}
In this experiment, we implemented the use of Auxiliary Rationale Memory (ARM) to establish our foundational ARM-RAG system. Correct examples from Experiment 5 were utilized to populate the Pyserini index. Both questions and answers, along with their reasoning chains—referred to as rationales—were indexed.

At query time, records retrieved from the index were used to construct prompts. Accuracy was measured separately for the training and test sets. It is important to note that the retrieval index contained examples from the training set but not from the test set. The results reflect this distinction: accuracy on the training set was 89.0\%, while accuracy on the test set was 75.3\%.

The superior performance on the training set can be attributed to the "hinting with the answer" effect, as described in Section \ref{subsec:exp2}. In most instances, the retriever fetches questions that are exact matches for the target question. In fact, analysis indicates that 78\% of the questions retrieved are exact matches for the target question.

The ARM-RAG system demonstrates a marginal improvement on the test set compared to the baseline, with an accuracy of 75.3\% for ARM-RAG versus 73.2\% for the non-ARM-RAG baseline. A detailed examination of the retrieval hits reveals that the system tends to retrieve examples that are superficially similar to the target question but overlooks the structural aspects of the problem. For instance, when presented with the target question shown in Figure \ref{fig:targetQuestion}, the system retrieves the examples displayed in Figure \ref{fig:targetQuestionHits}.

\begin{figure}[h]
\begin{small}
\begin{verbatim}
Ray buys a pack of hamburger meat for 
$5.00, a box of crackers for $3.50, 4 
bags of frozen vegetables at $2.00 
per bag and a pack of cheese for $3.50 
at the grocery store.  Because he is 
a store rewards member, he gets 10% 
off of his purchase.  What does his 
total grocery bill come to?
\end{verbatim}
\end{small}
\caption{Target Question}
\label{fig:targetQuestion}
\end{figure}

\begin{figure}[h]
\begin{small}
\begin{verbatim}
Frank goes to the store to buy some food. 
He buys 5 chocolate bars and 2 bags 
of chips. He hands the cashier $20 and 
gets $4 back as change. If the chocolate 
bars each cost $2, how much did each 
bag of chips cost?

In a grocery store, Julia bought 2 
pieces of Snickers and 3 packs of M&M's. 
If each piece of Snickers costs $1.5 and 
a pack of M&M's has the same cost as 2 
Snickers, how much is Julia's change 
if she gave the cashier 2 $10 bills?

Steve bought $25 worth of groceries. He 
bought a gallon of milk for $3, two 
boxes of cereal for $3.5 each, 4 bananas 
for $.25 each, four apples that cost $.5 
each and a number of boxes of cookies. 
The cookies cost twice as much per box 
as the gallon of milk. How many boxes of 
cookies did he get?
\end{verbatim}
\end{small}
\caption{Questions retrieved by target question}
\label{fig:targetQuestionHits}
\end{figure}

At a superficial level, the target question pertains to grocery shopping for food. The system retrieves questions related to grocery shopping for food but fails to consider the specific types of problems each question represents. For instance, the target question requires computation involving percentages, yet none of the retrieved examples involve problems that require understanding or calculating percentages. This indicates a limitation in the retrieval system's ability to discern and match the structural and conceptual aspects of the questions.

\subsection{Experiment 7: Obfuscated ARM-RAG}

As established in Section \ref{subsec:exp2}, prompting can significantly enhance the system's performance, while Section \ref{subsec:exp3} indicates that irrelevant prompts do not markedly affect performance. Experiment 6 seems to confirm these findings—the system retrieves examples that are superficially similar to the question being asked but lack structural resemblance to the problem within the question. Consequently, the prompts generated from these retrievals do not substantially aid the system. This underscores the importance of the retrieval system's ability to discern and match the deeper problem structure rather than surface-level similarities to improve the efficacy of the prompts.

In the subsequent experiment, we aimed to mitigate the influence of superficial similarity by obscuring the target question during the retrieval phase (but not during generation). This was achieved by replacing nouns with nonsensical words and proper names with very rare names. The identification of names and nouns was facilitated by GPT, using prompts such as "Give me all the male names in the following question...". This process resulted in a target query that is highly unlikely to match any entry in the KB. For instance, the target question depicted in Figure \ref{fig:targetQuestion} is transformed into an obfuscated version as shown in Figure \ref{fig:obfuscated}.

\begin{figure}[h]
\begin{small}
\begin{verbatim}
Halvard buys a plumbuzzle of spiggotwhap 
for $5.00, a dinglefrap of wobblegruff 
for $3.50, 4 crinklethorp of blibberfudge 
at $2.00 per bag and a plumbuzzle of 
trinkleshuff for $3.50 at the 
floopernoodle.  Because he is a 
snickerblast, he gets 10% off of his 
zibberflap.  What does his mumblestitch 
come to?
\end{verbatim}
\end{small}
\caption{Obfuscated target query}
\label{fig:obfuscated}
\end{figure}

Conducting the same experiment as Experiment 6, but utilizing obfuscated queries to search the knowledge base, resulted in an accuracy of 77.4\%. This represents an absolute improvement of 2.1\% over the system that did not use obfuscated queries.

The intention behind obfuscating queries was to prompt the system to retrieve answers that concentrate more on the structural aspects of a problem rather than superficial details. However, this strategy was only partially successful. Consider the 
examples retrieved by the query in Figure \ref{fig:obfuscated}, shown in Figure \ref{fig:obfuscatedHits}.

The results continue to revolve around the theme of shopping, even if not exclusively about food. This is probably because the obfuscated target query still contains the word "buy" and the unaltered word "bag," which suggests the purchase of small items. Somewhat promisingly, one of the retrieved questions involves percentages, which is a step towards the structural relevance that was absent in the non-obfuscated result set. Overall, obfuscation appears to exert a modestly positive influence on system performance, indicating that while it does not fully redirect the focus from superficial to structural aspects, it does make some progress in that direction.

\begin{figure}[h]
\begin{small}
\begin{verbatim}
Phillip's mother asked him to go to the 
supermarket to buy some things and gave 
him $95, so he spent $14 on oranges, $25 
on apples and $6 on candy. How much money 
does he have left?

Linda bought two coloring books at $4 
each, 4 packs of peanuts at $1.50 each 
pack, and one stuffed animal. She gave 
the cashier $25 and got no change. 
How much does a stuffed animal cost?

Carla bought 2 bags of mini peanut 
butter cups on clearance.  Each bag was 
$6.00 but was 75% off. How much did she 
spend on 2 bags of candy?
\end{verbatim}
\end{small}
\caption{Questions retrieved by obfuscated target query}
\label{fig:obfuscatedHits}
\end{figure}

\section{Analysis} 
Table \ref{table:data} summarizes the results of the main experiments.
\begin{table}[h!]
    \centering
\begin{center}
    \begin{tabular}{||c c||}
    \hline
    Experiment & Accuracy \\ [0.5ex]
    \hline\hline
    Baseline & 73.2\%  \\ 
    \hline
    ARM-RAG Test & 75.3\% \\
    \hline
    Obfuscated ARM-RAG Test & 77.4\% \\
    \hline
    \end{tabular}
\end{center}
\caption{Experiment results.}
\label{table:data}
\end{table}

Compared to the baseline system that does not utilize Retrieval Augmented Generation (RAG), the ARM-RAG system shows a slight improvement in performance. When obfuscation is applied to ARM-RAG, the impact on performance is more pronounced. The total increase in performance, when considering the obfuscated ARM-RAG system, is 5.7\% relative to the baseline. 

The strong-prompting technique highlighted in Section \ref{subsec:exp2} demonstrates that significant improvements in performance can be achieved through effective prompting. While ARM-RAG has shown some ability to enhance performance, the gains have not yet reached the potential maximum that seems possible. There is a likelihood that refining retrieval techniques could lead to further enhancements.

A promising direction for future research could involve developing methods to fully abstract the problem posed by a question or to classify it within a specific taxonomy. For instance, the target question shown in Figure \ref{fig:targetQuestion} could be categorized as a "discount of sum of products" problem.

\section{Conclusion} 

The central hypothesis of the paper is that Retrieval Augmented Generation (RAG) can be effectively utilized to enhance the problem-solving capabilities of Large Language Models (LLMs). The paper introduces ARM-RAG (Auxiliary Rationale Memory for Retrieval Augmented Generation), which employs Neural Information Retrieval to archive reasoning chains derived from solving grade-school math problems. A sequence of experiments demonstrates that the ARM-RAG system surpasses the performance of a baseline system that relies solely on LLMs.

\section*{Known Project Limitations}

This project has not been peer-reviewed.  There might be unknown bugs in
the code or other unintentional mistakes.

\section*{Authorship Statement}
This paper is the sole work of the author.

\bibliography{anthology,custom}

\end{document}